\documentclass[sn-mathphys-num,,Numbered]{sn-jnl}

\usepackage{lineno}
\usepackage{chngpage}

\usepackage{graphicx}
\graphicspath{ {Images/} }
\usepackage[caption=false]{subfig} 
\usepackage{float}
\usepackage{enumitem}

\usepackage{amsmath,amssymb,amsfonts} 

\usepackage{pgfplots}
\pgfplotsset{compat=newest}
\usepackage{tikz}

\usepackage{hyperref}
\hypersetup{colorlinks=true, citecolor=blue, urlcolor=black, linkcolor=magenta}
\usepackage{seqsplit}
\usepackage{xurl}
\usepackage{bbding} 

\usepackage{comment} 
\usepackage{verbatim}
\usepackage{minted} 
\usepackage{lmodern}

\usepackage{tabularx, tabularray, multirow}
\usepackage{xltabular}
\usepackage{array}
\newcolumntype{C}{>{\centering\arraybackslash}X}
\newcolumntype{L}{>{\raggedright\arraybackslash}X}
\newcolumntype{R}{>{\raggedright\arraybackslash}X}

\definecolor{blueaccent}{RGB}{0,150,214}
\definecolor{greenaccent}{RGB}{0,139,43}
\definecolor{purpleaccent}{RGB}{130,41,128}
\definecolor{orangeaccent}{RGB}{240,83,50}
\definecolor{palesilver}{rgb}{0.96, 0.96, 0.96}
\definecolor{pastelgray}{rgb}{0.81, 0.81, 0.77}
\definecolor{arsenic}{rgb}{0.23, 0.27, 0.29}

\begin{document}

\title{Securing Self-supervised Data Curation for Foundation Models Robustness}
\author*[1,2]{\fnm{Sandeep} \sur{Gupta}}\email{s.gupta@qub.ac.uk}
\author[2]{\fnm{Roberto} \sur{Passerone}}\email{roberto.passerone@unitn.it}

\affil*[1]{\orgdiv{Centre for Secure Information Technologies (CSIT)}, \orgname{Queen's University Belfast}, \orgaddress{\country{UK}}}
\affil[2]{\orgdiv{Department of Information Engineering and Computer Science}, \orgname{University of Trento}, \orgaddress{\country{Italy}}}

\abstract{Self-supervised data curation provides a pathway to scaling and improving the generalization capabilities of machine learning models. By leveraging self-supervised learning (SSL) for data curation, the demand for massive training datasets required by foundation models can be effectively met. SSL greatly alleviates the costs associated with annotation and manual dataset curation while minimizing the need for human oversight. However, the integrity of SSL-curated datasets must be rigorously checked, as reliance on anonymous and unvetted external sources can substantially increase the risk of data poisoning. In this paper, we propose a Poisoned Data Detector (PDD), an active defense mechanism designed to ensure the integrity of SSL-curated datasets prior to foundation model training. PDDs are designed using a combination of the pretrained ImageBind model and traditional classifiers, including Random Forest (RF), k-Nearest Neighbors (KNN), Naive Bayes (NB), and Support Vector Machines (SVM). We rigorously evaluated PDDs using 176,200 images from three diverse datasets and three different adversarial attacks encompassing both in-distribution and out-of-distribution scenarios. Notably, SVM-PDD achieves superior performance for both in-distribution (Set3–Set5) and out-of-distribution (TrueFace and 140K RealFace) datasets. Our design demonstrates strong scalability and enables the rapid integration of new adversarial attack detectors through an ensemble approach.
}

\keywords{Foundation models, self-supervised learning, data curation, poisoning attack}
\maketitle

\section{Introduction}
Visual Foundation Models (VFMs) have shown the capability to effectively acquire visual features, producing representations that exhibit strong generalization and transferability. These representations integrate both image and text features within a shared semantic space, facilitating their application to a diverse array of downstream tasks in a zero-shot or one-shot manner~\cite{liu2024few}. VFMs can be pre-trained models typically trained on vast datasets comprising billions of image-text pairs. The training process is efficiently automated through self-supervised data curation using models such as Contrastive Language-Image Pre-training (CLIP)~\cite{radford2021learning} and Deeper Into Neural Networks (DINOv2)~\cite{oquab2024dinov}. Additionally, large language models (LLMs) enhance the generation of training data by producing images from text prompts. Self-supervised learning (SSL) is a subset of unsupervised learning that extracts discriminative features from unlabeled data to aid in labeling and automate data curation~\cite{gui2024survey}. SSL supports various tasks, including segmentation, classification, and regression, offering a scalable and cost-effective alternative to the resource-intensive and time-consuming process of manual data labeling. 

At present, the prevalent real-world SSL curation pipelines used to produce billion-scale datasets for visual foundation models include the CLIP-driven web-scraping approach of LAION~\cite{schuhmann2022laion}, the benchmark-driven filtering framework of DataComp~\cite{li2024datacomp}, and the embedding-based retrieval pipeline employed in DINOv2~\cite{oquab2024dinov}. Table~\ref{tab:Pipelines} compares the three pipelines, each illustrating the shift toward automated, scalable curation in SSL. Although they aim to reduce dependence on labeled data and mitigate noise in web-scale sources, their reliance on large volumes of unvetted external data introduces multiple avenues for potential poisoning attacks.
\begin{table*}[!ht]
    \centering
    \hyphenpenalty 10000
    \footnotesize
    \caption{A comparison between the prevalent real-world SSL curation pipelines}\label{tab:Pipelines}
    \begin{tblr}{width=1\linewidth,
        colspec = {p{.09\linewidth} p{.26\linewidth} p{.26\linewidth} p{.26\linewidth} },
    }\hline
    \textbf{Aspect} & \textbf{LAION}~\cite{schuhmann2022laion} & \textbf{DataComp}~\cite{li2024datacomp} & \textbf{DINOv2 Retrieval}~\cite{oquab2024dinov}\\\hline
    Data Modality & Image-text pairs & Image-text pairs & Images only\\\hline
    Core Focus & Web scraping + CLIP filtering & Benchmark for curation strategies & Embedding-based retrieval from web\\\hline
    Scale & 5.85B pairs & 12.8B candidates; subsets up to 1B & 1.2B candidates; 142M images\\\hline
    Key Innovation & Distributed common crawl parsing + safety scores & Fixed models for fair data comparison & Self-supervised kNN expansion\\\hline
    Output Use & CLIP/Stable Diffusion training & Leaderboard-evaluated CLIP models & ViT self-distillation (e.g., DINOv2)\\\hline
    \end{tblr}
\end{table*}

\begin{figure}[!ht]
    \centering
    \footnotesize
    \includegraphics[width=1\linewidth]{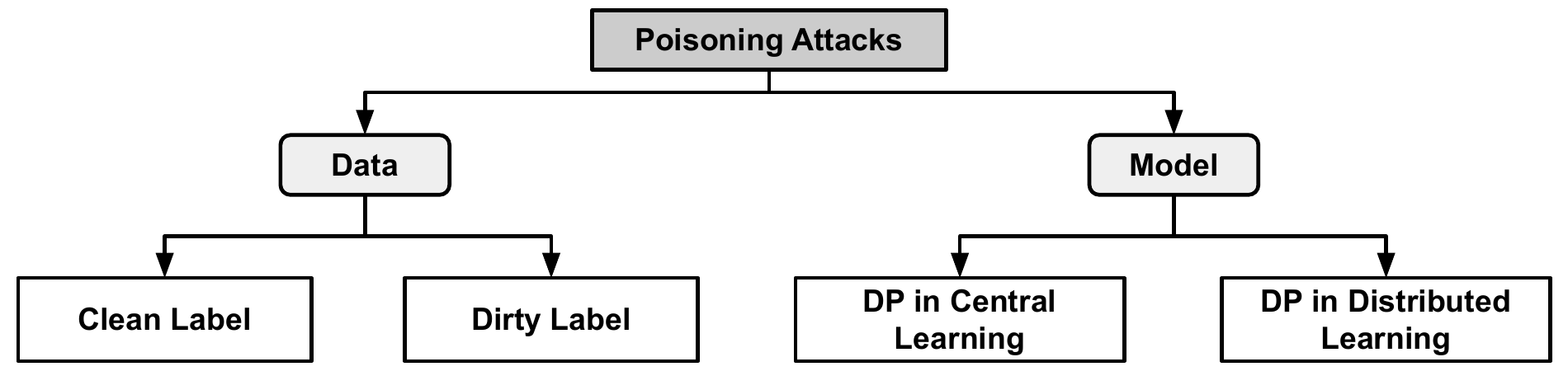}
    \caption{Illustration of poisoning attacks on data and models. Data poisoning (DP) attacks involve tampering with training data, such as dirty-label or clean-label data poisoning. Model poisoning attacks directly compromise the model itself, depending on whether the victim model is trained in a centralized or distributed manner.}\label{fig:PoisoningAttackTaxonomy}
\end{figure}

As VFMs expand in data and model dimensionality, SSL is emerging as a critical research focus~\cite{oquab2024dinov}. However, poisoning attacks pose a significant threat to VFMs training, as adversaries can manipulate training data or disrupt the training process to maliciously alter the model behavior~\cite{wang2022threats}. As illustrated in Figure~\ref{fig:PoisoningAttackTaxonomy}, \textit{Data poisoning attacks} involve tampering with training data, whereas \textit{model poisoning attacks} directly compromise the model itself. Data poisoning attacks can be further classified as dirty-label or clean-label. Dirty-label attacks manipulate both the data content and labels, while clean-label attacks alter only the data content. Pre-labeled datasets or data from anonymous and unverified sources can serve as potential entry points for data poisoning attacks, indirectly compromising the integrity of the final system. Malicious third-party platforms, including Machine Learning as a Service (MLaaS) providers and distributors of pre-trained models and codebases, can introduce model poisoning attacks by intentionally manipulating the target system. 

Defense strategies against poisoning attacks can be broadly categorized as \emph{passive}, which focus on post-attack remediation, or \emph{active}, which implement preventive measures~\cite{wang2022poisoning}. In this paper, we propose a novel mechanism for detecting poisoned data in datasets curated using SSL for VFM training, which can be described as an active defense mechanism. Malicious actors can exploit SSL pipelines to orchestrate data poisoning attacks and upload tainted datasets online. This vulnerability poses a substantial security risk, particularly in environments where comprehensive monitoring of VFM training is infeasible. Given the potential for significant downstream impact, ensuring the integrity of the training data is as crucial as securing the final VFM deployment.

\noindent
The main contributions of this paper are highlighted below.
\begin{itemize}[leftmargin=*]
    \item Propose a Poisoned Data Detector (PDD) to verify SSL-curated datasets integrity before they are used for foundation model training as an active defense mechanism. Our design exhibits strong scalability and facilitates the rapid integration of new adversarial attack detectors through an ensemble approach.
    \item PDDs are designed using a combination of the pretrained ImageBind model and traditional classifiers, including Random Forest (RF), k-Nearest Neighbors (KNN), Naive Bayes (NB), and Support Vector Machines (SVM). Each PDD is trained on 10\%, 20\%, 30\%, 40\%, and 50\% of the data, yielding twenty PDD variants in total.
    \item Rigorous experiments are performed to evaluate twenty PDDs using 176,200 images from three diverse datasets and three different adversarial attacks encompassing both in-distribution (ID) and out-of-distribution (OOD) scenarios. SVM-PDD demonstrates superior performance in detecting poisoned data.
\end{itemize}
\vspace{2mm}

The rest of the paper is structured as follows: Section~\ref{sec:Background} provides an overview of foundation models, self-supervised data curation, and real-world self-supervised data curation pipelines. Section~\ref{sec:Problem} describes data poisoning attacks on data curated using SSL to subvert a VFM during its training phase. Section~\ref{sec:Methodology} defines the methodology for designing the proposed solution, including network and dataset selection, poisoned dataset generation, and the poisoned data detector. Section~\ref{sec:ED} presents the high-level design of the proposed solution for SSL-curated dataset integrity checks and the experimental results. Section~\ref{sec:relatedwork} discusses the related work to data poisoning attacks. Finally, Section~\ref{sec:Conclusions} concludes the article and outlines future directions.

\section{Background}\label{sec:Background}
Visual Foundation Models (VFMs) are large-scale pre-trained models that learn rich visual representations from massive image-text or image-only datasets, enabling strong generalization across diverse downstream tasks such as classification, detection, segmentation, and pose estimation in zero- or few-shot settings~\cite{liu2024few, gupta2025investigation}.

\subsection{Self-supervised data curation}
\begin{figure}[!ht]
    \centering
    \includegraphics[width=0.9\linewidth, keepaspectratio]{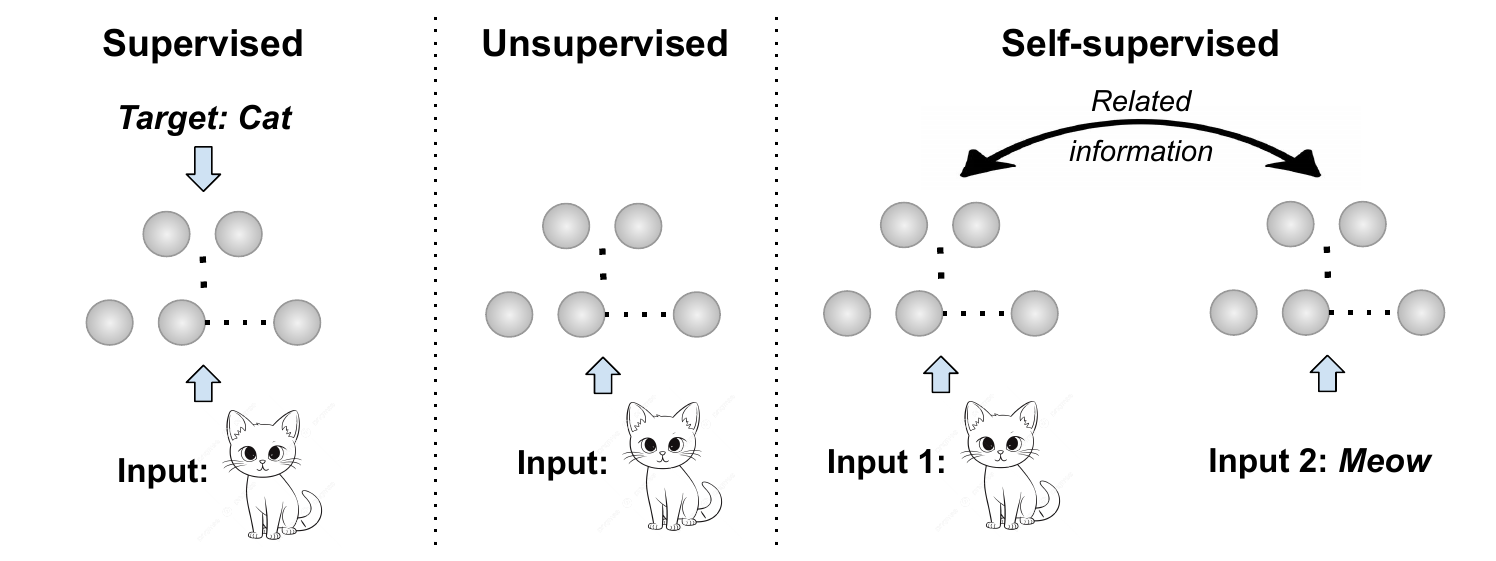}
    \caption{Illustration of supervised learning, unsupervised learning, and SSL. SSL can automatically generates a large volume of labeled data for visual foundation models training addressing the limitations of manual labeling process.}\label{fig:SSL}
\end{figure}

Figure~\ref{fig:SSL} depicts a comparison between supervised-, unsupervised-, and self-supervised learning. Supervised learning typically requires large volumes of manually labeled data to train models effectively and achieve superior performance~\cite{gui2024survey}. In contrast, unsupervised learning discovers patterns in unlabeled data without explicit supervision. Given the cost and time associated with manual annotation, self-supervised data curation has emerged as an efficient strategy for generating labeled data, reducing reliance on human labels and supporting the pretraining of foundation models. 
Recent self-supervised methods have attracted considerable attention for their ability to exploit structure in unlabeled data and improve representation learning, helping alleviate some limitations of manual labeling~\cite{liu2021self}. Self-supervised approaches are commonly grouped into generative, contrastive, and adversarial families, although exact taxonomies vary across surveys. Generative SSL trains an encoder to map input data ($x$) to a latent representation ($z$), and a decoder to reconstruct $x$ from $z$. Contrastive SSL learns representations by maximizing similarity between different views of the same data ($x$) in the latent space ($z$). Generative-contrastive SSL employs an encoder-decoder to generate synthetic data, and a discriminator to distinguish it from real data, thereby learning robust representations.

Methods such as CLIP~\cite{radford2021learning} and DINOv2~\cite{oquab2024dinov} efficiently automate the VFM training process. CLIP~\cite{radford2021learning} matches captions to images as a simple pre-training task. CLIP utilizes contrastive learning to jointly train image and text encoders, matching images to their corresponding text descriptions by distinguishing them from other (image, text) pairs within a batch. The method successfully yielded 400 million image-text pairs from the internet, demonstrating its scalability and effectiveness in generating massive datasets in an SSL setting. Like the GPT family, CLIP demonstrates emergent capabilities during pre-training, excelling in tasks such as OCR, geo-localization, and action recognition. DINOv2~\cite{oquab2024dinov} automates the creation of tailored, diverse, and curated image datasets, offering a significant improvement over traditional uncurated approaches. DINOv2 is trained using various Vision Transformer (ViT) architectures on a dataset of 142 million images. Subsequently, the curated pre-training dataset is constructed by retrieving images from the uncurated data source that closely resemble those in the curated sources including ImageNet-1k and Google Landmarks. The process involves computing high-dimensional image embeddings using a self-supervised ViT-H/16 model pre-trained on ImageNet-22k. These embeddings are then used to calculate cosine similarity, enabling effective clustering of the uncurated data with k-means. SSL can serve as an alternative to text-guided pre-training, supporting tasks such as segmentation, classification, and regression. Moreover, LLMs can facilitate image generation from text prompts, further expanding the training data pool. 

\subsection{Real-world self-supervised data curation pipelines}
While CLIP and DINOv2 established the foundational principles of automated data curation, contemporary foundation-model training relies on substantially larger and more sophisticated pipelines operating at web scale. We outline the three dominant paradigms currently employed to construct billion-scale datasets without requiring any manual labeling.

\subsubsection{LAION-style Pipelines}
As described in Table~\ref{tab:LAION}, the LAION pipeline is a three-stage, distributed web-scraping and filtering framework designed to construct large-scale, open-source, CLIP-filtered image–text datasets~\cite{schuhmann2022laion}. It begins with massive web archives such as Common Crawl (CC) and produces high-quality image-text pairs (e.g., 5.85 billion in LAION-5B). The pipeline emphasizes aesthetic and semantic alignment without manual curation, relying on CLIP embeddings for automated quality scoring.

\begin{table*}[!ht]
    \centering
    \hyphenpenalty 10000
    \footnotesize
    \caption{LAION high-level architecture stages}\label{tab:LAION}
    \begin{tblr}{width=1\linewidth,
        colspec = {p{.08\linewidth} p{.2\linewidth} p{.28\linewidth} p{.31\linewidth} },
    }\hline
    \textbf{Stage} & \textbf{Description} & \textbf{Key Components/Tools} & \textbf{Data Flow/Input-Output}\\\hline
    \SetCell[r=2]{l}{Candidate Extraction} & \SetCell[r=2]{l}{Parse web archives to identify image-text pairs (e.g., $<$img alt=``..."$>$ tags).} & \SetCell[r=2]{l}{Distributed cluster (e.g., AWS workers)~\cite{de2024openwhisk}; HTML parsing on CC snapshots ($\approx$300 TiB/month, $\approx$3B pages).} & Input: Raw CC HTML files. \\\hline
    & & & Output: $\approx$20-30B candidate URL-text pairs (unfiltered).\\\hline
    \SetCell[r=2]{l}{Download and Initial Processing} &  \SetCell[r=2]{l}{Download images and compute basic metadata (e.g., resolution, deduplication).} & \SetCell[r=2]{l}{img2dataset (parallel downloader)~\cite{beaumont2021img2dataset}; PySpark for deduplication; GPU nodes for fast early filtering.} & Input: Candidate URLs.\\\hline
    & & & Output: Downloaded images and text ($\approx$12-15B pairs); remove broken/invalid samples.\\\hline
    \SetCell[r=2]{l}{Quality Filtering and Embedding} & \SetCell[r=2]{l}{Score pairs for semantic alignment (image-text similarity) and safety.} &  \SetCell[r=2]{l}{CLIP, ViT-B/32 or ViT-L/14 for embeddings; cosine similarity threshold (e.g., 0.28 for English); Autofaiss for kNN indexing~\cite{douze2025faiss}; NSFW/watermark detectors~\cite{kastryulin2025yaart}.} & Input: Processed pairs. \\\hline
    & & & Output: Filtered dataset (e.g., 5.85B pairs in LAION-5B: 2.3B English and 2.2B multilingual, 1B unlabeled); includes embeddings and safety scores.\\\hline
    \end{tblr}
\end{table*}

\subsubsection{DataComp Benchmark Pipelines}
As described in Table~\ref{tab:DataComp}, DataComp is a benchmark framework for evaluating data-curation strategies rather than a single dataset~\cite{li2024datacomp}. 
\begin{table*}[!ht]
    \centering
    \hyphenpenalty 10000
    \footnotesize
    \caption{DataComp high-level architecture stages}\label{tab:DataComp}
    \begin{tblr}{width=1\linewidth,
        colspec = {p{.1\linewidth} p{.26\linewidth} p{.25\linewidth} p{.26\linewidth} },
    }\hline
    \textbf{Stage} & \textbf{Description} & \textbf{Key Components/Tools} & \textbf{Data Flow/Input-Output}\\\hline
    \SetCell[r=2]{l}{Candidate Pool Generation} & \SetCell[r=2]{l}{Extract raw image-text pairs from web sources.} & \SetCell[r=2]{l}{Common Crawl parsing; initial deduplication (e.g., exact/near-duplicates via hashing)~\cite{wenzek2020ccnet}.} & Input: CC archives ($\approx$240T tokens total).\\\hline
    & & & Output: 12.8B uncurated image-text pairs (candidate pool).\\\hline
    \SetCell[r=2]{l}{Curation and Filtering} & \SetCell[r=2]{l}{Apply participant-defined strategies (e.g., quality scoring, mixing). Baselines: CLIP similarity (top 0.3 fraction) and image-based filters (e.g., aesthetic, watermark detection)~\cite{capasso2024comprehensive}.} & \SetCell[r=2]{l}{OpenCLIP for scoring~\cite{ilharco2021openclip}; deduplication (rust-based for efficiency); data mixing (e.g., balance classes). YAML configs for Ray-based pipelines.} & Input: Candidate pool.\\\hline
    & & & Output: Curated subset (e.g., DataComp-1B baseline: 1B high-quality pairs).\\\hline
    \SetCell[r=2]{l}{Standardized Training} & \SetCell[r=2]{l}{Evaluation,Train fixed CLIP model on curated data; test on downstream tasks.} & \SetCell[r=2]{l}{OpenLM framework~\cite{ilharco2023openlm}; scales: XS/S/M/L/XL; 53 evaluations (e.g., ImageNet zero-shot, segmentation).} & Input: Curated dataset.\\\hline
    & & & Output: Model performance metrics (e.g., 79.2\% ImageNet accuracy for DataComp-1B, +3.7\% over OpenAI CLIP).\\\hline
    \end{tblr}
\end{table*}
It standardizes the model architecture (e.g., CLIP) and training hyperparameters, requiring participants to innovate specifically in data selection and filtering from a shared candidate pool. The framework conducts evaluation through a fixed training pipeline and comprehensive downstream testing (typically across 38-53 tasks). It provides two tracks: Filtering, which permits the use only of the provided candidate pool, and BYOD (`bring your own data'), which allows the incorporation of external data in addition to the pool. DataComp spans multiple scales, ranging from XS (approximately 412M parameters, $\approx$4 GPU-hours) to XL (approximately 7B parameters, $\approx$40k GPU-hours).
    
\subsubsection{DINOv2 Retrieval Pipeline (LVD-142M)}
As described in Table~\ref{tab:DINOv2}, the DINOv2 pipeline is a fully automated, self-supervised system for curating large-scale, image-only datasets without relying on text annotations~\cite{oquab2024dinov}. Starting from a small set of high-quality `seed' images drawn from curated sources such as ImageNet, the pipeline retrieves visually similar examples from massive uncurated web collections. These images are embedded using a self-supervised ViT-H/16 model and matched via cosine similarity, enabling efficient clustering and large-scale expansion of the curated set. The result is a balanced and diverse dataset—exemplified by LVD-142M (142M images) that supports high-quality pre-training through self-distillation using DINO and iBOT losses. By grounding the selection process in robust visual embeddings rather than labels, the pipeline ensures scalable dataset growth while maintaining semantic coherence and visual quality.

\begin{table*}[!ht]
    \centering
    \hyphenpenalty 10000
    \footnotesize
    \caption{DINOv2 high-level architecture stages}\label{tab:DINOv2}
    \begin{tblr}{width=1\linewidth,
        colspec = {p{.11\linewidth} p{.18\linewidth} p{.26\linewidth} p{.31\linewidth} },
    }\hline
    \textbf{Stage} & \textbf{Description} & \textbf{Key Components/Tools} & \textbf{Data Flow/Input-Output}\\\hline
    \SetCell[r=2]{l}{Embedding Generation} & \SetCell[r=2]{l}{Extract features from curated and uncurated images.} & \SetCell[r=2]{l}{Pre-trained ViT-H/14 (self-supervised on ImageNet-22k); high-dimensional embeddings (e.g., 1024-dim).} & Input: Curated seeds ($\approx$1-10M images from ImageNet-22k, Google Landmarks) + uncurated pool (~1.2B web images).\\\hline
    & & & Output: Embeddings for all images.\\\hline
    \SetCell[r=2]{l}{Deduplication} & \SetCell[r=2]{l}{Remove near-duplicates to increase diversity} & \SetCell[r=2]{l}{Copy detection (e.g., perceptual hashing); filter out test/validation set overlaps.} & Input: Embeddings and images.\\\hline
    & & & Output: Deduplicated uncurated set ($\approx$reduced by 10-20\%).\\\hline
    \SetCell[r=2]{l}{Retrieval} & \SetCell[r=2]{l}{Augmentation, Match uncurated images to seeds via similarity; Balance concepts.} & \SetCell[r=2]{l}{Cosine similarity and k-means clustering; kNN retrieval; self-supervised expansion.} & Input: Deduplicated embeddings.\\\hline
    & & & Output: Augmented dataset (LVD-142M: 142M diverse images, balanced across concepts).\\\hline
    \end{tblr}
\end{table*}

While these automated pipelines enable unprecedented scale, their dependence on external, unverified data sources can introduce critical risks of data poisoning.

\section{Related work}\label{sec:relatedwork}
Defense mechanisms against poisoning attacks can be divided into passive defenses, which remediate attacks after they occur, and active defenses, which prevent attacks proactively~\cite{wang2022poisoning}. Training data sanitization employs filtering algorithms to identify, remove, or revise poisoned data points within the training dataset. Model parameter sanitization focuses on detecting anomalies in the victim model parameters and subsequently correcting them to restore normal functionality. Active defense mitigates data poisoning by employing data preprocessing techniques, including cleaning, enhancement, and transformation, to scrutinize raw training data. Robust training methods further strengthen defenses by making the training process resilient to poisoning attacks. Countermeasures against poisoning attacks include training data filtering, robust learning, and auxiliary tools~\cite{wang2022poisoning}. Training data filtering, employing techniques such as input manipulation detection and gradient shaping, mitigates the impact of outliers and offers attack-agnostic robustness. Model verification and robustification techniques enhance adversarial resilience against model poisoning attacks and effectively detect backdoor attacks. Integrating auxiliary tools like Generative Adversarial Networks (GANs) and robust statistical methods can significantly improve model classification capabilities by refining learned representations and emphasizing critical signal components.

Fares and Nandakumar~\cite{fares2024attack} propose Attack To Defend (A2D) approach that detects poisoned models by exploiting their inherent susceptibility to adversarial perturbations. The authors introduce Sensitivity to Adversarial Perturbations (SAP) to quantify a model sensitivity to adversarial attacks at a specific perturbation bound. The metric enables a normalized comparison of models adversarial sensitivity (robustness). A poisoned target model exhibits a higher SAP than a benign model, which can be leveraged for detecting poisoned models. Ishmam and Thomas~\cite{ishmam2024semantic} propose semantic shield for defending against backdoor and poisoning attacks on contrastively trained vision language models by enforcing knowledge-guided train-time constraints. This method integrates a prompting technique using an open-source language model to freely extract constituent knowledge elements from any caption. 

Chen et al.~\cite{chen2021pois} present De-Pois, an attack-agnostic defense designed to mitigate diverse poisoning attacks, including Targeted Clean-Label Poisoning, Poisoning Attack With GAN, Label-Flipping Attack, and Regression Attack. The approach involves training a companion model on clean samples to approximate the target model behavior. Poisoned samples are identified by detecting significant discrepancies between the predictions of the companion and target models, eliminating the need for specific knowledge about the attack methodology or underlying machine learning algorithms. 
Tejankar et al.~\cite{tejankar2023defending} propose PatchSearch, a method that defends SSL models against patch-based data poisoning attacks by identifying and filtering out poisoned samples. This approach follows a three-stage pipeline: (1) initial model training on a potentially poisoned dataset, (2) identification and removal of poisoned samples using the PatchSearch algorithm, and (3) final model training on the cleaned dataset. Guan et al.~\cite{guan2022few} propose Shapley Pruning framework that leverages neuron interaction analysis to identify and mitigate backdoor attacks in poisoned models. This method prunes infected neurons, preserving model structure and accuracy. 

Zhang et al.~\cite{zhang2023backdoor} propose Causality-inspired Backdoor Defense (CBD) that learns deconfounded representations for classification. This approach employs causal graph analysis to model backdoor attacks as confounding variables, which introduce spurious correlations between input images and target labels. Specifically, the backdoor trigger is conceptualized as a confounder, establishing a non-causal path exploited by the network during training. This path leads to erroneous predictions when the trigger is present, effectively highlighting the vulnerability targeted by backdoor attacks. Peri et al.~\cite{peri2020deep} propose Deep k-NN, a representation-based defense that detects clean-label poisoning by computing the average distance of each sample to its k-nearest neighbors in the feature space of a lightly fine-tuned proxy model. Poisoned samples within the target class exhibit significantly larger distances than clean samples, enabling outlier removal without requiring a separate clean reference set. Table~\ref{tab:ComparisonDP} provides a comparison of existing defenses against data poisoning, with a focus on key aspects relevant to SSL-curated datasets for foundation models.

{
\centering
\footnotesize
\hyphenpenalty 10000
\begin{xltabular}{1\linewidth}{p{.1\linewidth} p{.15\linewidth} p{.14\linewidth} p{.14\linewidth} p{.14\linewidth} p{.14\linewidth}}
    \caption{Comparison of Data Poisoning Defenses}\label{tab:ComparisonDP}\\\hline
    \textbf{Defense} & \textbf{Core Mechanism} & \textbf{Threat Model} & \textbf{Strengths} & \textbf{Limitations} & \textbf{Relevance to SSL Curation}\\\hline
    Activation Clustering, 2018~\cite{chen2023tutorial} & Clusters neuron activations in the target class's penultimate layer (using k-means, k=2); removes the smaller (anomalous) cluster as poisoned. & Backdoor attacks where poisoned samples form distinct activation clusters from clean ones in the target class. & Simple, no clean data needed; near-100\% F1-score on MNIST/CIFAR-10 backdoors; fast post-training. & Assumes clear separation in activation space (fails on distributed or clean-label poisons); not robust to adaptive attacks that mimic clean activations. & Useful for SSL (e.g., DINOv2 embeddings) but requires a preliminary model; untested on massive unlabeled sets.\\\hline
    Fine-Pruning, 2018~\cite{liu2018fine} & Prunes suspicious neurons (identified via activation analysis) and fine-tunes the model on remaining clean data to excise backdoors. & Backdoor poisoning via data injection; targets DNNs like CNNs/ResNets. & Retains high clean accuracy post-defense; effective against targeted triggers; lightweight pruning. & Needs access to the full model; less effective on over-parameterized foundation models; vulnerable if poison embeds deeply. & Applicable to VFM pre-training but may disrupt SSL representations; scales poorly to billion-scale data without acceleration.\\\hline
    Deep k-NN, 2021~\cite{peri2020deep} & k-NN distance in deep representation space of proxy model; high-distance samples are considered poison & Clean-label backdoor & SOTA on clean-label at time; no clean reference needed; works on representations & Requires held-out set for proxy; $O(n^2)$ k-NN search (mitigable) & Embeddings could be used on unlabeled web data for fast outlier detection.\\\hline
    SPECTRE, 2021~\cite{hayase2021spectre} & Robust covariance estimation on intermediate representations to amplify spectral signatures of poisoned data, enabling backdoor removal via retraining. & Backdoor attacks (e.g., trigger-based) on distributed training data; assumes small poisoning fraction ($<$1\%). & Highly effective at low poisoning rates; removes backdoors entirely without trusted data; strong theoretical guarantees. & Relies on detectable spectral signatures (may fail on subtle clean-label poisons); compute-intensive for very large datasets. & Could filter unvetted web-scraped data in LAION-style pipelines by scanning embeddings pre-training.\\\hline
    De-Pois, 2021~\cite{chen2021pois} & Trains a mimic model (via GAN-augmented clean data simulation) to replicate clean behavior; filters poisons by prediction divergence from the target model. & General data poisoning (label-flip, clean-label, backdoor); attack-agnostic. & Broad applicability ($>$90\% F1 on multiple attacks/datasets); uses GANs for data augmentation without full clean set. & GAN training overhead; vulnerable to adaptive poisons that align mimic divergence; assumes some clean data access. & Ideal for SSL (augments unlabeled data); could integrate with ImageBind-like embeddings for pre-training checks.\\\hline
    Shapley Pruning, 2022~\cite{guan2022few} & Neuron interaction pruning via Shapley values & Backdoor in poisoned models & Preserves structure/ accuracy; few-shot effective & Computationally intensive for large models & Applicable to post-curation model hardening.\\\hline
    CBD, 2023~\cite{zhang2023backdoor} & Causal graph deconfounding for representations & Backdoor via spurious correlations & Models confounders explicitly; improves generalization & Causal graph complexity; training overhead & Enhances SSL feature robustness against triggers.\\\hline
    PatchSearch, 2023~\cite{tejankar2023defending} & Iterative clustering of image patches using SSL model features; scores and prunes clusters likely containing patch-based poisons. & Patch-based backdoors in SSL (e.g., localized triggers in unlabeled images); no trusted data assumed. & Tailored to SSL vulnerabilities; modular for large-scale filtering. & Limited to visible patch triggers (not subtle perturbations like PGD); requires initial SSL training on suspect data. & Directly addresses patch poisons in web-scale SSL curation (e.g., LAION); ensemble-friendly like the paper's PDD.\\\hline
    Semantic Shield, 2024~\cite{ishmam2024semantic} & Knowledge-guided train-time constraints via prompting & Backdoor/ poisoning in contrastive VLMs & Integrates LLMs for caption analysis; attack-agnostic & Computationally heavy prompting; caption-dependent & Directly targets CLIP-like SSL pipelines.\\\hline
    A2D, 2024~\cite{fares2024attack} & Sensitivity to adversarial perturbations (SAP metric) & Poisoned models (adversarial susceptibility) & Normalized robustness comparison; no attack knowledge needed & Relies on perturbation bounds; less effective on subtle poisons & Can be useful for post-SSL encoder verification.\\\hline
\end{xltabular}
}

\section{Problem description}\label{sec:Problem}
\begin{figure}[!ht]
    \centering
    \includegraphics[width=0.98\linewidth, keepaspectratio]{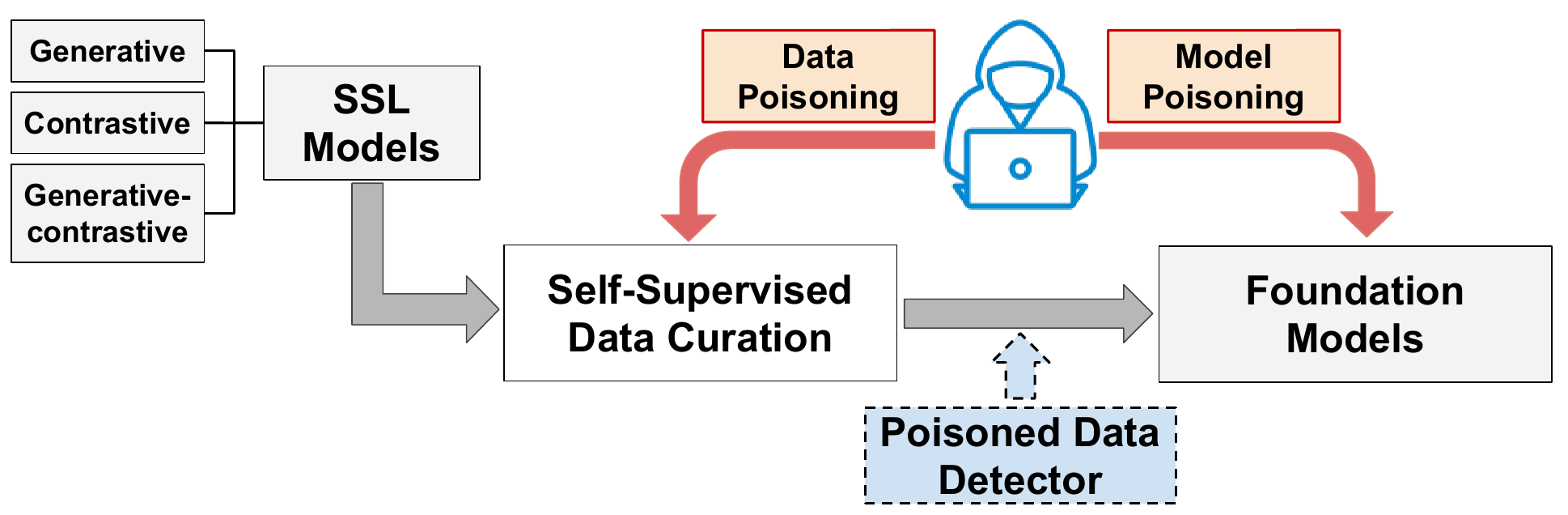}
    \caption{Illustration of a simplified training framework for foundation models, including poisoning attacks and data poisoning detector.}\label{fig:FoundationModel}
\end{figure}

Figure~\ref{fig:FoundationModel} highlights the critical risks posed by data poisoning attacks to the SSL data curation pipeline, emphasizing how these attacks can compromise the integrity of the training process by introducing malicious data, ultimately leading to intentional manipulation of the VFM~\cite{wang2022threats}. Adversaries can execute poisoning attacks by strategically injecting, deleting, or altering training data, including adversarial examples, with the intent to corrupt the training data and produce defective, biased, or unreliable VFMs~\cite{zhou2022adversarial}. In particular, we investigate data poisoning attacks targeting data curated using SSL, which can manipulate the VFM under development.

Within environments where full oversight of VFM training is impractical, poisoning attacks pose a notable threat~\cite{wang2022poisoning}. Consequently, securing the training phase is just as crucial as ensuring the integrity of the final VFM deployment. In particular, scenarios in which SSL-curated datasets are sourced externally from anonymous and unvetted sources present a significant risk of both untargeted and targeted data poisoning by malicious actors. Adversaries performing untargeted data poisoning strive for a denial of service (DoS) effect by increasing the overall prediction error rate. In contrast, targeted data poisoning attacks aim to manipulate the VFM predictions for a particular class.


\section{Methodology}\label{sec:Methodology}

\subsection{Networks}
We select the ConvNeXT\footnote{\url{https://github.com/facebookresearch/ConvNeXt}} model, trained on a dataset of 1K classes with 1.3M images, due to its superior performance across a wide range of vision tasks, including ImageNet\footnote{\url{https://www.image-net.org/}} classification, object detection and segmentation on COCO\footnote{\url{https://cocodataset.org}}, and semantic segmentation on ADE20K\footnote{\url{https://ade20k.csail.mit.edu/}}~\cite{liu2022convnet}. ConvNeXt is a modernized pure convolutional architecture that achieves Transformer-level performance by adopting several design principles inspired by Vision Transformers and Swin Transformers (e.g., large convolution kernels, inverted bottlenecks, LayerScale, and fewer normalization layers) while remaining entirely convolution-based and containing no self-attention mechanisms. We additionally use a standard Convolutional Neural Network (CNN) as a reference model when generating poisoned datasets to evaluate the proposed solution. CNNs are a subclass of feedforward neural networks specifically designed to extract hierarchical spatial features through learned convolutional filters~\cite{li2021survey}.

\subsection{Datasets}\label{sec:Datasets}
\begin{figure*}[!ht]
    \centering
    \captionsetup{format=hang,font=small, margin=5pt}
    \hyphenpenalty 10000
    \subfloat[ConvNeXT achieves an 95.71\% accuracy on Set1]{
        \includegraphics[width=1\linewidth, keepaspectratio]{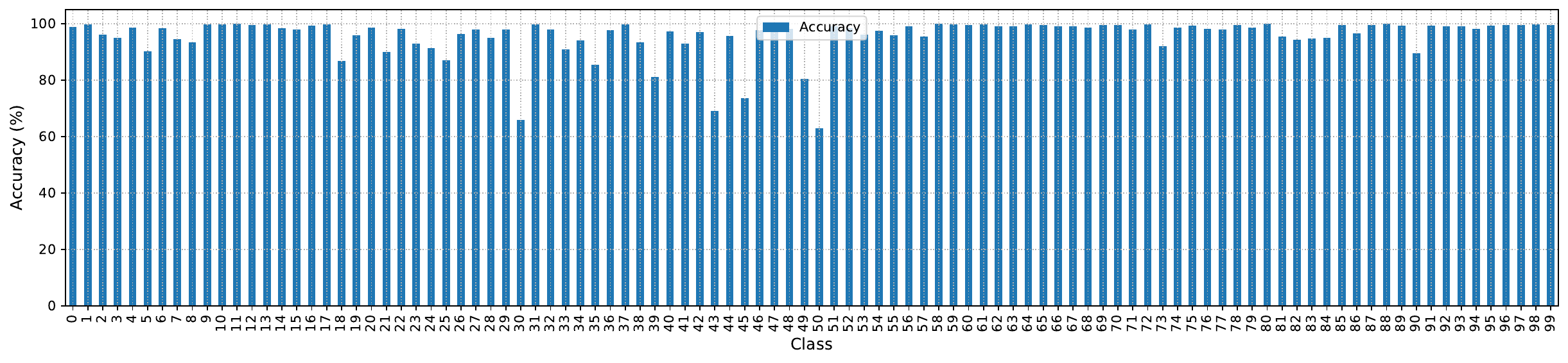}
    }\\
    \subfloat[ConvNeXT achieves an 88\% accuracy on Set2]{
        \includegraphics[width=1\linewidth, keepaspectratio]{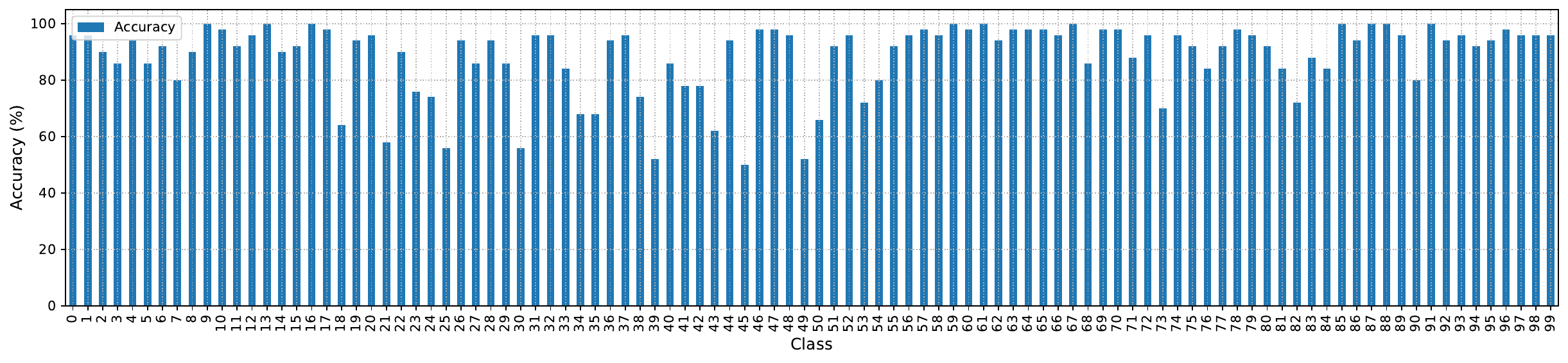}
    }
    \caption{The ConvNeXT model, with a total of 197.74 million parameters, performance on a 100-class subset of the ImageNet dataset.}\label{fig:ConvNeXTAcc}
\end{figure*}

We utilize a subset of the ImageNet dataset (Set1, Set2) with 100 classes for our experiments. Table~\ref{tab:Dataset} provides details of the various datasets constructed from Set1 and Set2. The model achieves an accuracy of 95.71\% on Set1 and 88\% on Set2, and the class-wise accuracy is illustrated in Figure~\ref{fig:ConvNeXTAcc}. Subsequently, we use Set3 and Set4, which contain 10 classes and achieve 99.57\% and 100\% classification accuracy respectively, when tested with the ConvNeXT pretrained model, to design a poison data detector.
\begin{table*}[!ht]
    \centering
    \hyphenpenalty 10000
    \footnotesize
    \caption{Datasets description leverage in the experiments and ConvNeXT classification performance}\label{tab:Dataset}
    \begin{tblr}{width=1\linewidth,
        colspec = {p{.1\linewidth} p{.23\linewidth} p{.07\linewidth} p{.32\linewidth} p{.1\linewidth}},
    }\hline
    \SetCell[r=2]{l}{\textbf{Dataset}} & \SetCell[r=2]{l}{\textbf{Source}} & \SetCell[r=2]{l}{\textbf{Classes}} & \SetCell[c=2]{c}{\textbf{ConvNeXT classification performance}}\\\hline
    & & & \textbf{Dataset specifications} & \textbf{Accuracy}\\\hline
    \SetCell[r=2]{l}{ImageNet100} & \SetCell[r=2]{l}{\href{https://www.kaggle.com/datasets/ambityga/imagenet100}{Kaggle}} & \SetCell[r=2]{l}{100} & Set1: 130000 (\textit{1300 images per class}) & 95.71\% \\\hline
    & & & Set2: 5000 (\textit{50 images per class}) & 88.00\%\\\hline
    \SetCell[r=2]{l}{ImageNet10 Normal} & \SetCell[r=2]{l}{Subsets selected from ImageNet100} & \SetCell[r=2]{l}{10} & Set3: 13000 (\textit{1300 images per class}) & 99.57\% \\\hline
    & & & Set4: 500 (\textit{50 images per class}) & 100.00\% \\\hline
    \SetCell[r=2]{l}{ImageNet10 Poisoned} & \SetCell[r=2]{l}{Generated using Projected Gradient Descent (PGD) attack} & \SetCell[r=2]{l}{10} & Set5: 13000 (\textit{1300 images per class}) & 8.45\%\\\hline
    & & & Set6: 500 (\textit{50 images per class}) &  9.20\%\\\hline
    \end{tblr}
\end{table*}

We select the Trueface PostSocial~\cite{boato2022trueface} and 140K-Real-and-Fake-Faces~\cite{kaggle2025RealFake} datasets to evaluate the proposed solution for poisoned data detection. The Trueface PostSocial and 140K-Real-and-Fake-Faces datasets consist of 2,500 and 70,000 real images, respectively. The generation of poisoned images for each dataset is carried out using the Projected Gradient Descent (PGD)~\cite{madry2018towards} attack with a four-layer CNN model containing 102.89 million parameters. Subsequently, both normal and poisoned embeddings are generated using ImageBind~\cite{girdhar2023imagebind}. In addition, we generate a small poisoned dataset (100 samples) for different attacks, such as the Fast Gradient Sign Method (FGSM)~\cite{goodfellow2015explaining} and Carlini and Wagner (C\&W)~\cite{carlini2017towards} using Set3, to evaluate the efficacy of PDDs against diverse attacks.

\subsection{Poisoned dataset generation}\label{sec:PoisonedDSGen}
We use gradient-based method that directly manipulates input data using model loss gradients to generate a poisoned dataset. To determine the most effective method for generating poisoned data, we evaluate the robustness of the ConvNeXT model against various techniques, including adversarial attacks crafted using the FGSM and PGD. The FGSM attack uses the gradients of the loss function with respect to the input data to determine the direction in which the input should be modified to maximize the model's error. The attack operates in three steps: First, it calculates the loss after a forward pass. Second, it computes the gradient of this loss with respect to the image pixels. Finally, it subtly alters the image pixels in the direction that maximizes the loss. PGD is an iterative method for generating adversarial examples. It works by repeatedly taking a small step in the direction that maximizes the model prediction error (using the FGSM) and then projecting the result back onto the space of valid inputs. 

\begin{figure*}[!ht]
    \centering
    \captionsetup{format=hang,font=small, margin=5pt}
    \hyphenpenalty 10000
    \subfloat[Performance against the FGSM attack on Set3. Accuracy degrades from 100\% to 91.16\%.]{
        \includegraphics[width=0.48\linewidth, keepaspectratio]{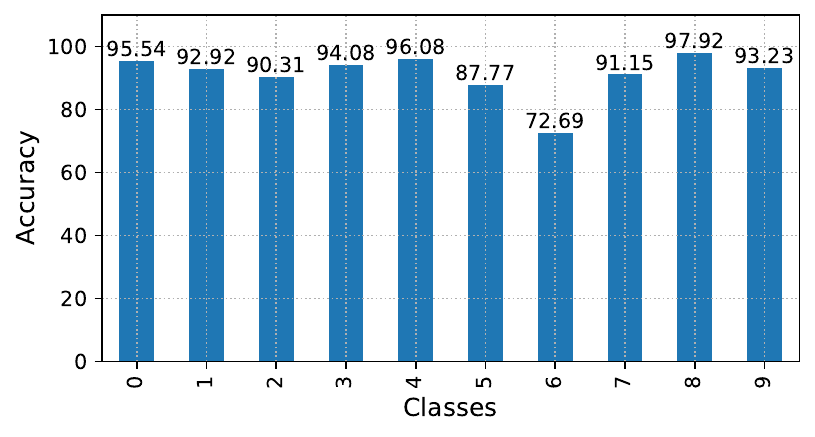}
    }
    \subfloat[Performance against the PGD attack on Set3. Accuracy degrades to 8.4\% from 100 \%]{
        \includegraphics[width=0.48\linewidth, keepaspectratio]{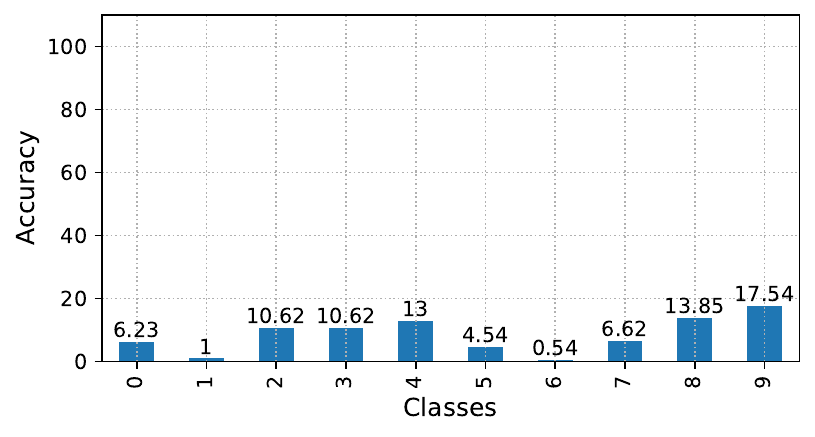}
    }
    \caption{ConvNeXT model performance against FGSM and PGD attacks.}\label{fig:ConvNeXTAdv}
\end{figure*}

Figure~\ref{fig:ConvNeXTAdv} compares the performance of the ConvNeXT model against the FGSM and PGD attacks with a perturbation radius of $0.1$ using $\ell_{\infty}$ on Set3. The optimal value for the perturbation radius is determined empirically. It can be observed that PGD, a multi-step variant of the FGSM algorithm, is more effective at deceiving the model. Therefore, we constructed a 10-class poisoned dataset: Set5 (13,000 images) and Set6 (500 images), using PGD-based adversarial perturbations.

\subsection{Poisoned data detection}
\begin{figure}[!ht]
    \centering
    \captionsetup{format=hang,font=small, margin=5pt}
    \hyphenpenalty 10000
    \subfloat[Set3 (Normal) vs. Set5 (Poisoned)\label{fig:DSSet3Set5}]{
        \includegraphics[width=0.48\linewidth, keepaspectratio]{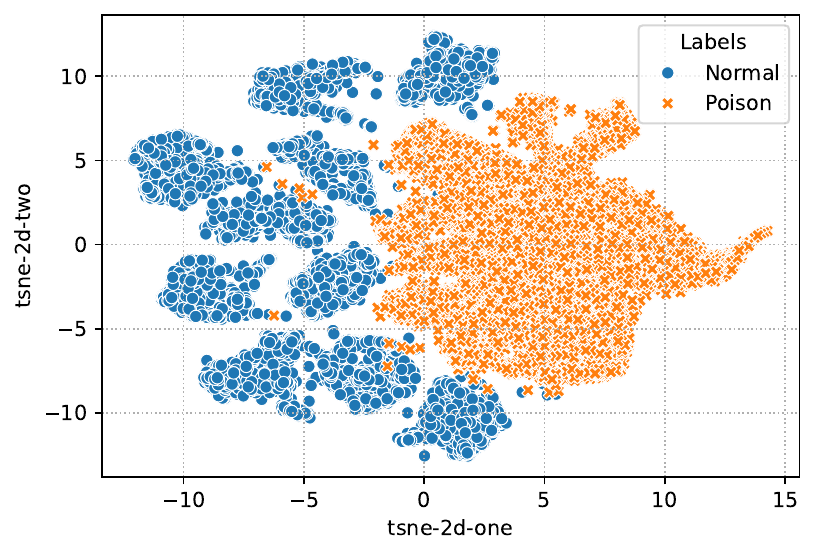}
    }
    \subfloat[Set4 (Normal) vs. Set6 (Poisoned)\label{fig:DSSet4Set6}]{
        \includegraphics[width=0.48\linewidth, keepaspectratio]{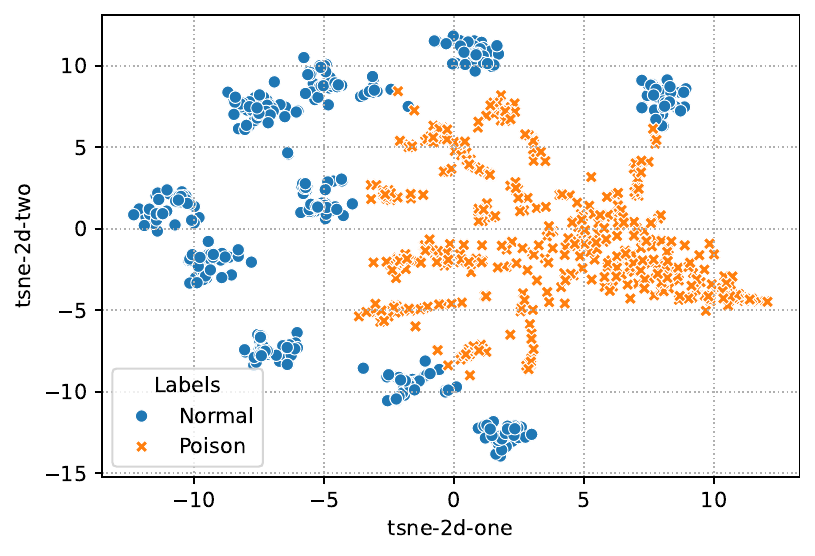}
    } \\
    \subfloat[Set4 (Normal class-wise data distribution)\label{fig:DSSet4}]{
        \includegraphics[width=0.48\linewidth, keepaspectratio]{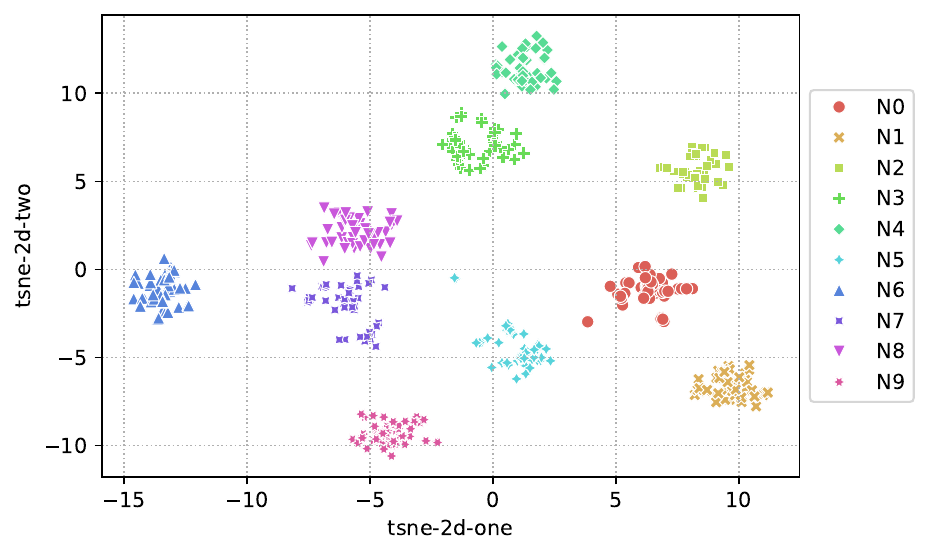}
    }
    \subfloat[Set6 (Poisoned class-wise data distribution). \label{fig:DSSet6}]{
        \includegraphics[width=0.48\linewidth, keepaspectratio]{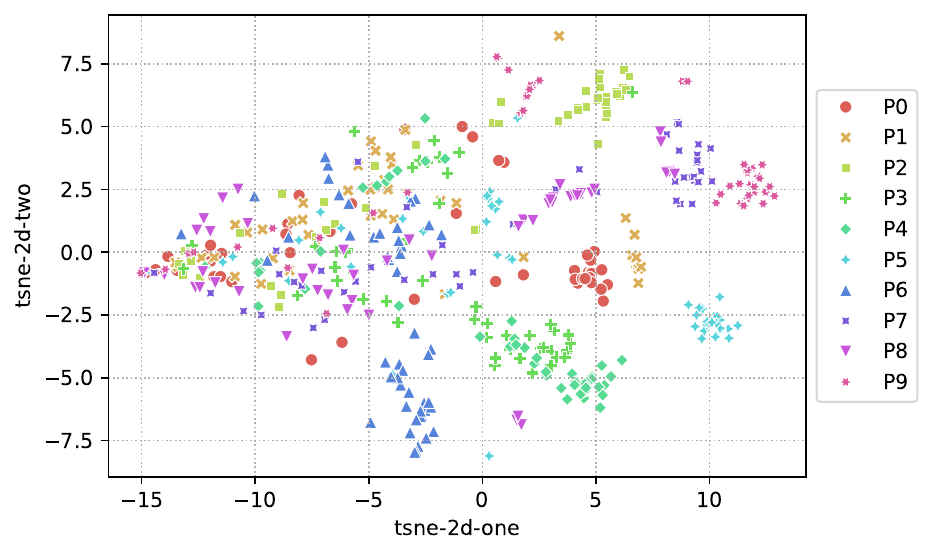}
    }
    \caption{A comparison of the embeddings generated by ImageBind model for the normal and poisoned datasets using t-SNE plots.}\label{fig:DatasetComparison}
\end{figure}

As illustrated in Figure~\ref{fig:FoundationModel}, we exploit a pretrained ImageBind model\footnote{\url{https://github.com/facebookresearch/ImageBind}}, which has 1.2 billion parameters, to extract embeddings from both the normal and generated poisoned datasets~\cite{girdhar2023imagebind}. ImageBind can also provide a single
joint embedding ($E_{\mathcal{I},\mathcal{M}}$), where where $\mathcal{I}$
represents images and $\mathcal{M}$ are other modalities, viz., text, audio, thermal depth, and Inertial Measurement Units (IMUs). Figure~\ref{fig:DatasetComparison} compares the embeddings generated by the ImageBind model for the normal and poisoned datasets by plotting their Kullback-Leibler (KL) divergence, i.e., relative entropy, between the probability distributions in both high-dimensional and low-dimensional spaces. These low-dimensional representations are generated using the T-Distributed Stochastic Neighbor Embedding (t-SNE) algorithm~\cite{van2008visualizing}. 

The t-SNE dimensionality reduction technique is designed to visualize high-dimensional data and assess the separability of two classes. It fits and transforms high-dimensional vectors into a space with a reduced number of dimensions, typically two or three, for better comparison through 2-D or 3-D plots. Figure~\ref{fig:DSSet3Set5} analyzes conditional probabilities across 26,000 samples of Set3 and Set5, a KL divergence of 3.12 is achieved after 300 iterations. Similarly, Figure~\ref{fig:DSSet4Set6} analyzes conditional probabilities across 500 samples of Set4 and Set6, a KL divergence of 0.71 is achieved after 300 iterations. Figure~\ref{fig:DSSet4} and Figure~\ref{fig:DSSet6} compare the class-wise data distribution of Set4 (Normal) and Set6 (Poisoned), respectively. The result demonstrates the ImageBind model ability to effectively separate normal and poisoned image features, indicating a clear distinction between the two categories.

\section{Experimental details}\label{sec:ED}
\begin{figure}[!ht]
    \centering
    \includegraphics[width=.9\linewidth, keepaspectratio]{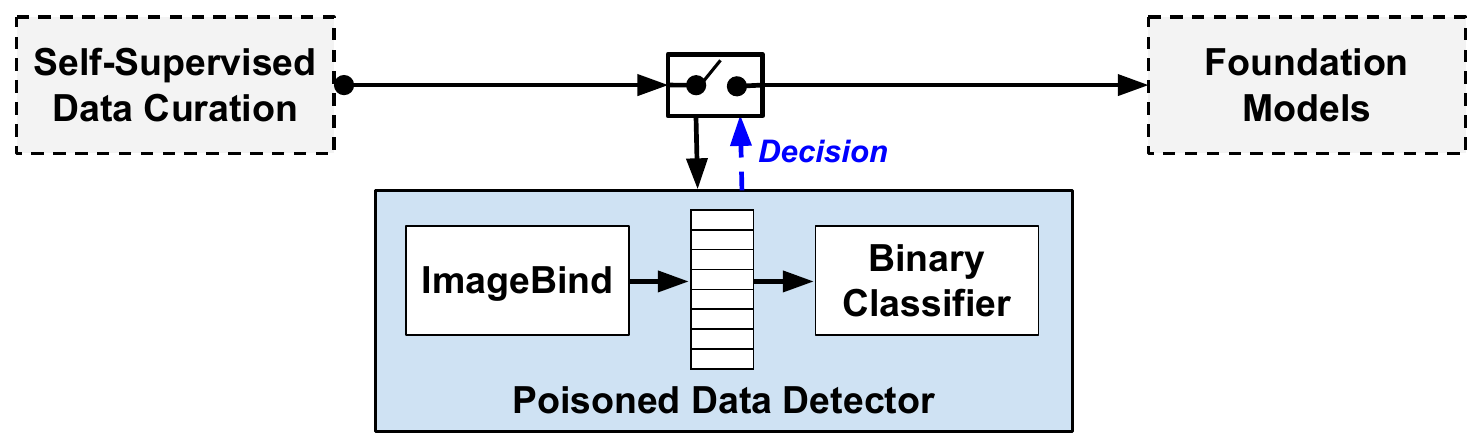}
    \caption{A high-level design Poisoned data detector (PDD).}\label{fig:PoisonedDataDetector}
\end{figure}

Figure~\ref{fig:PoisonedDataDetector} presents the high-level design of the proposed PDD, enabling the verification of SSL curated datasets prior to it is used for foundation model training.

\subsection{Binary classifiers}
We employ diverse classification algorithms, including Random Forest (RF), k-Nearest Neighbors (KNN), Naïve Bayes (NB), and Support Vector Machines (SVM), to design binary classifiers that distinguish between normal and poisoned images. Table~\ref{tab:Set46Result} presents the performance (i.e., accuracy) of each classifier trained on Set4 (500 normal images) and Set6 (500 poisoned images) using the holdout testing method. The evaluation process involves dividing the combined dataset of 1,000 images into two non-overlapping subsets for training and testing, with the training size varying from 10\% to 50\% (with step size of 10\%). Subsequently, we evaluate five variants of four classifiers on three diverse datasets, one of which is ID and two are OOD.
\begin{table*}[!ht]
    \centering
    \hyphenpenalty 10000
    \footnotesize
    \caption{PDDs are trained on Set4 and Set6 with training sizes ranging from 10\% to 50\%, and performance is reported in terms of accuracy (\%).}\label{tab:Set46Result}
    \begin{tblr}{width=1\linewidth,
        colspec = {p{.08\linewidth} p{.42\linewidth} p{.05\linewidth} p{.05\linewidth} p{.05\linewidth} p{.05\linewidth} p{.05\linewidth}},
    }\hline
    \SetCell[r=3]{c}{\textbf{Classifier}} & \SetCell[r=3]{c}{\textbf{Hyperparameters}} & \SetCell[c=5]{c}{\textbf{Accuracy (\%)}}\\\hline
    & & \SetCell[c=5]{c}{\textbf{Training percentage}}\\\hline
    & & \textit{10\%} & \textit{20\%} & \textit{30\%} & \textit{40\%} & \textit{50\%}\\\hline
    SVM & `C': 1, `gamma': 1, `kernel': `rbf' & 100.00 & 100.00 & 100.00 & 100.00 & 100.00\\\hline
    RF & `bootstrap': True, `max\_depth': 50, `max\_features': `sqrt', `min\_samples\_leaf': 2, `min\_samples\_split': 10, `n\_estimators': 30 & 98.22 & 98.50 & 98.71 & 98.83 &  99.40\\\hline
    NB & `priors': None, `var\_smoothing': 0.657933224657568 & 98.67 & 99.25 & 99.14 & 99.17 & 99.00 \\\hline
    KNN & `metric': `minkowski', `n\_neighbors': 5, `weights': `uniform' & 96.67 &  99.00 & 99.57 & 100.0 & 100.0\\\hline
    \end{tblr}
\end{table*}

\begin{figure}[!ht]
    \centering
    \subfloat[SVM]{
    \includegraphics[width=0.48\linewidth]{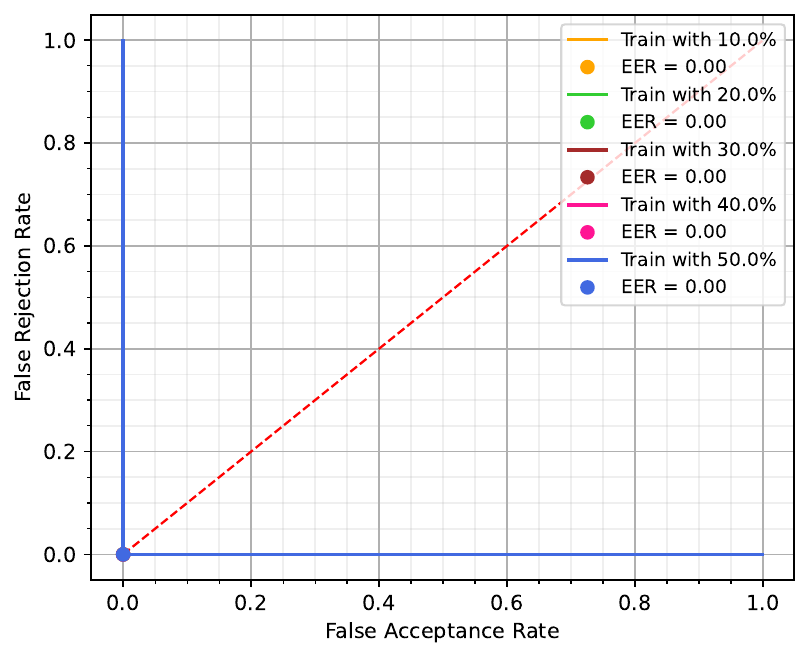}
    }
    \subfloat[RF]{
    \includegraphics[width=0.48\linewidth]{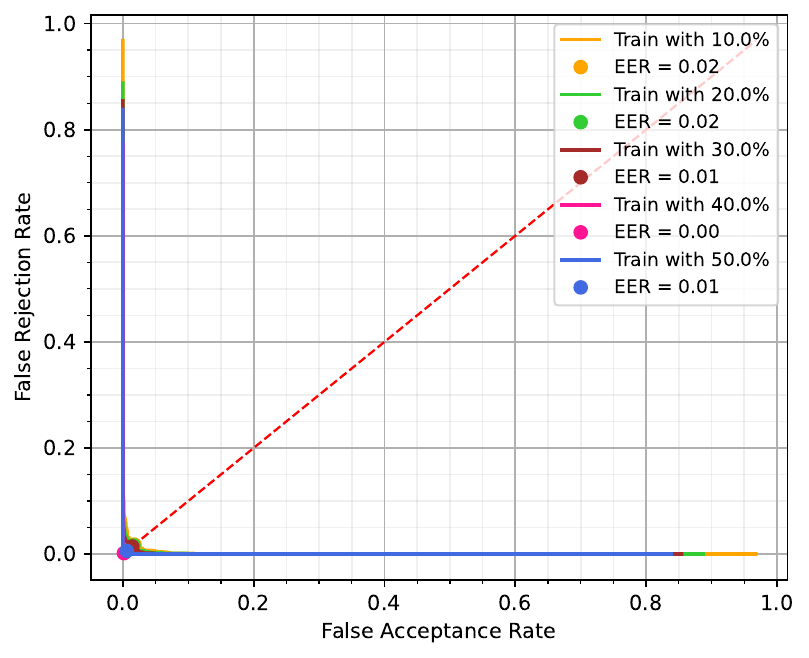}
    }\\
    \subfloat[NB]{
    \includegraphics[width=0.48\linewidth]{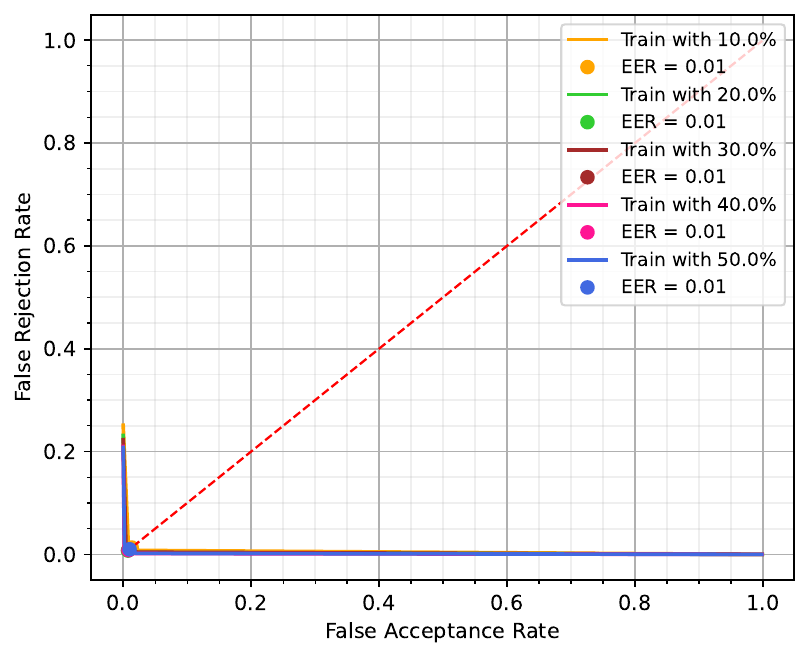}
    }
    \subfloat[KNN]{
    \includegraphics[width=0.48\linewidth]{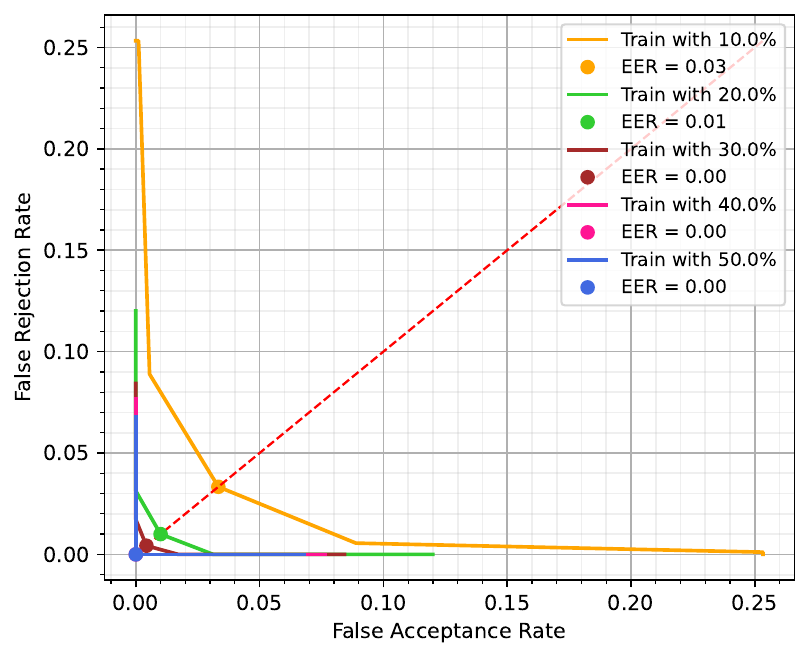}
    }
    \caption{DET plots for the PDDs designed using SVM, RF, NB, and KNN with varying training percentage from 10\% to 50\% of the total Set4-Set6 dataset.}\label{fig:DET}
\end{figure}

We exploit grid search for hyperparameter optimization, systematically exploring a predefined range of values for each selected classifier hyperparameters to determine the optimal settings. This method efficiently identifies the best-performing combination of hyperparameter values for the chosen classifier~\cite{bischl2023hyperparameter}. Each classifier is trained and evaluated across all grid combinations to identify the top-performing configuration that can deliver superior performance. It can be observed that the SVM-based PDDs outperforms others, primarily due to its effectiveness in handling high-dimensional datasets and scenarios where the number of features surpasses the number of samples~\cite{gupta2024visual}. Figure~\ref{fig:DET} presents the detection error tradeoff (DET) plots of the PDDs designed using SVM, RF, NB, and KNN, with varying training percentages from 10\% to 50\% (with step size of 10\%) of the total samples in the Set4-Set6 dataset.

\subsection{Evaluation results}
\subsubsection{Using different dataset}
\begin{figure}[!ht]
    \centering
    \footnotesize
    \subfloat[PDDs performance on the Set3 and Set5 combination, containing 13,000 normal and 13,000 poisoned images.\label{fig:Set35Result}]{
    \begin{tikzpicture}[scale=1]
    \begin{axis}[
        style={
            /pgf/number format/fixed,
            /pgf/number format/fixed zerofill,
            /pgf/number format/precision=2
        },
        width=0.9\linewidth,
        height=3cm,
        ylabel= Accuracy (\%),
        xtick={0,1,2,3,4,5},
        xticklabels={,PDD-10\%, PDD-20\%, PDD-30\%, PDD-40\%, PDD-50\%},
        ymajorgrids,
        grid=both,
    	legend style = {at={(0.5, 1.1)},
    					anchor=south,legend columns=5},
        ]
        \addplot[color=greenaccent,solid, mark=o, mark options={solid}] coordinates { (1, 99.95) (2, 99.94) (3, 99.96) (4, 99.97) (5, 99.97) };
        \addplot[color=purpleaccent,solid, mark=+, mark options={solid}] coordinates { (1, 98.72) (2, 98.78) (3, 99.15 ) (4, 99.54) (5, 99.63)};
        \addplot[color=orangeaccent, solid, mark=triangle*, mark options={solid}] coordinates { (1, 99.04) (2, 99.61) (3, 99.48) (4, 99.62) (5, 99.60)};
        \addplot[color=blueaccent,solid, mark=x, mark options={solid}] coordinates { (1, 97.06) (2, 98.30) (3, 99.27) (4, 99.72) (5, 99.76)};
        
        \legend{SVM, RF, NB, KNN}
    \end{axis}
    \end{tikzpicture}
    }\\
    \subfloat[PDDs performance on the TrueFace dataset containing 5,000 normal and 5,000 poisoned images.\label{fig:TrueFaceResult}]{
    \begin{tikzpicture}[scale=1]
    \begin{axis}[
        style={
            /pgf/number format/fixed,
            /pgf/number format/fixed zerofill,
            /pgf/number format/precision=2
        },
        width=0.9\linewidth,
        height=3cm,
        ylabel= Accuracy (\%),
        xtick={0,1,2,3,4,5},
        xticklabels={,PDD-10\%, PDD-20\%, PDD-30\%, PDD-40\%, PDD-50\%},
        ymajorgrids,
        grid=both,
    	legend style = {at={(0.5, 1.1)},
    					anchor=south,legend columns=5},
        ]
        \addplot[color=greenaccent,solid, mark=o, mark options={solid}] coordinates { (1, 99.95) (2, 99.94) (3, 99.96) (4, 99.97) (5, 99.78) };
        \addplot[color=purpleaccent,solid, mark=+, mark options={solid}] coordinates { (1, 82.6) (2, 71.44) (3, 77.84) (4, 90.84) (5, 93.16)};
        \addplot[color=orangeaccent, solid, mark=triangle*, mark options={solid}] coordinates { (1, 82.52) (2, 78.84) (3, 75.84) (4, 75.66) (5, 76.66)};
        \addplot[color=blueaccent,solid, mark=x, mark options={solid}] coordinates { (1, 87.72) (2, 84.86) (3, 83.34) (4, 83.36) (5, 84.04)};
    \end{axis}
    \end{tikzpicture}
    }\\
    \subfloat[PDDs performance on the 140k RealFace dataset containing 70,000 normal and 70,000 poisoned images.\label{fig:140KRealFaceResult}]{
    \begin{tikzpicture}[scale=1]
    \begin{axis}[
        style={
            /pgf/number format/fixed,
            /pgf/number format/fixed zerofill,
            /pgf/number format/precision=2
        },
        width=0.9\linewidth,
        height=3cm,
        ylabel= Accuracy (\%),
        xtick={0,1,2,3,4,5},
        xticklabels={,PDD-10\%, PDD-20\%, PDD-30\%, PDD-40\%, PDD-50\%},
        ymajorgrids,
        grid=both,
    	legend style = {at={(0.5, 1.1)},
    					anchor=south,legend columns=5},
        ]
        \addplot[color=greenaccent,solid, mark=o, mark options={solid}] coordinates { (1, 99.48) (2, 99.81) (3, 99.80) (4, 99.85) (5, 99.86) };
        \addplot[color=purpleaccent,solid, mark=+, mark options={solid}] coordinates { (1, 79.49) (2, 75.39) (3, 78.96) (4, 90.11) (5, 91.61)};
        \addplot[color=orangeaccent, solid, mark=triangle*, mark options={solid}] coordinates { (1, 78.29) (2, 76.64) (3, 75.02) (4, 74.73) (5, 76.54)};
        \addplot[color=blueaccent,solid, mark=x, mark options={solid}] coordinates { (1, 79.69) (2, 76.25) (3, 75.37) (4, 76.11) (5, 77.99)};
    \end{axis}
    \end{tikzpicture}
    }
    \caption{A comparison of PDDs performance using different datasets. Note: different scaling is used in above figures for better readability.}\label{fig:PDDdatasets}
\end{figure}
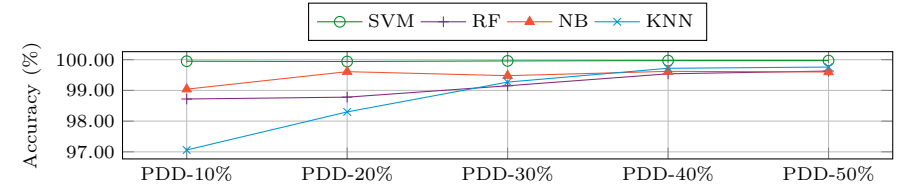
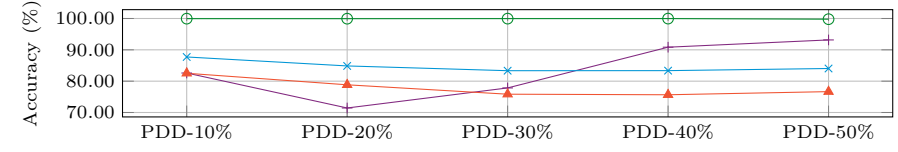
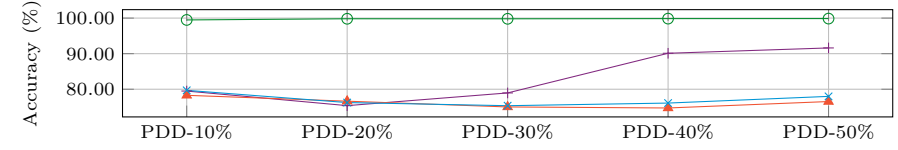

We evaluate all twenty PDDs using three distinct datasets. Figure~\ref{fig:Set35Result},~\ref{fig:TrueFaceResult}, and~\ref{fig:140KRealFaceResult} present the performance of each PDD on the Set3–Set5, TrueFace postsocial dataset~\cite{boato2022trueface}, and 140K RealFace datasets~\cite{kaggle2025RealFake}, respectively. SVM-PDD achieves superior performance for ID dataset, i.e., Set3–Set5 and for OOD datasets, i.e., TrueFace and 140K RealFace. 

\subsubsection{Against different attacks}
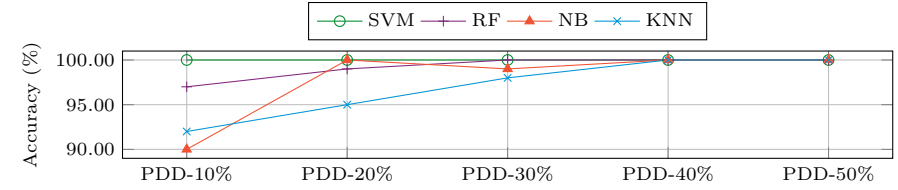
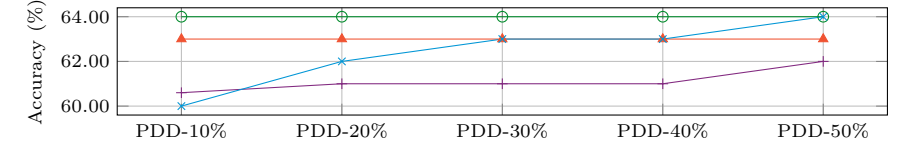
\begin{figure}[!ht]
    \centering
    \footnotesize
    \subfloat[PDDs performance against FGSM-poisoned images.\label{fig:Set3FGSMResult}]{
    \begin{tikzpicture}[scale=1]
    \begin{axis}[
        style={
            /pgf/number format/fixed,
            /pgf/number format/fixed zerofill,
            /pgf/number format/precision=2
        },
        width=0.9\linewidth,
        height=3cm,
        ylabel= Accuracy (\%),
        xtick={0,1,2,3,4,5},
        xticklabels={,PDD-10\%, PDD-20\%, PDD-30\%, PDD-40\%, PDD-50\%},
        ymajorgrids,
        grid=both,
    	legend style = {at={(0.5, 1.1)},
    					anchor=south,legend columns=5},
        ]
        \addplot[color=greenaccent,solid, mark=o, mark options={solid}] coordinates { (1, 100.0) (2, 100.0) (3, 100.0) (4, 100.0) (5, 100.0) };
        \addplot[color=purpleaccent,solid, mark=+, mark options={solid}] coordinates { (1, 97.0) (2, 99.0) (3, 100.0) (4, 100.0) (5, 100.0)};
        \addplot[color=orangeaccent, solid, mark=triangle*, mark options={solid}] coordinates { (1, 90.0) (2, 100.0) (3, 99.0) (4, 100.0) (5, 100.0)};
        \addplot[color=blueaccent,solid, mark=x, mark options={solid}] coordinates { (1, 92.0) (2, 95.0) (3, 98.0) (4, 100.0) (5, 100.0)};
        \legend{SVM, RF, NB, KNN}
    \end{axis}
    \end{tikzpicture}
    }\\
    \subfloat[PDDs performance against C\&W-poisoned data.\label{fig:Set3CWResult}]{
    \begin{tikzpicture}[scale=1]
    \begin{axis}[
        style={
            /pgf/number format/fixed,
            /pgf/number format/fixed zerofill,
            /pgf/number format/precision=2
        },
        width=0.9\linewidth,
        height=3cm,
        ylabel= Accuracy (\%),
        xtick={0,1,2,3,4,5},
        xticklabels={,PDD-10\%, PDD-20\%, PDD-30\%, PDD-40\%, PDD-50\%},
        ymajorgrids,
        grid=both,
    	legend style = {at={(0.5, 1.1)},
    					anchor=south,legend columns=5},
        ]
        \addplot[color=greenaccent,solid, mark=o, mark options={solid}] coordinates { (1, 64.0) (2, 64.0) (3, 64.0) (4, 64.0) (5, 64.0) };
        \addplot[color=purpleaccent,solid, mark=+, mark options={solid}] coordinates { (1, 60.6) (2, 61.0) (3, 61.0) (4, 61.0) (5, 62.0)};
        \addplot[color=orangeaccent, solid, mark=triangle*, mark options={solid}] coordinates { (1, 63.0) (2, 63.0) (3, 63.0) (4, 63.0) (5, 63.0)};
        \addplot[color=blueaccent,solid, mark=x, mark options={solid}] coordinates { (1, 60.0) (2, 62.0) (3, 63.0) (4, 63.0) (5, 64.0)};
    \end{axis}
    \end{tikzpicture}
    }
    \caption{A comparison of PDDs performance against different attacks using only poisoned data. Note: different scaling is used in above figures for better readability.}\label{fig:PDDAttacks}
\end{figure}

Considering the practical challenges of defending against clean-label data poisoning attacks, we also evaluate PDDs against FGSM and C\&W attacks to assess their performance on OOD data. In this evaluation, each dataset contains only 100 poisoned samples from the respective attack, as explained in Section~\ref{sec:Datasets}. The performance of PDDs against these two types of poisoned data is reported in Figure~\ref{fig:Set3FGSMResult} and~\ref{fig:Set3CWResult}. PDDs have shown good performance to detect FGSM-poisoned data, however, performance in detecting C\&W-poisoned data is relatively lower. Studies~\cite{rottmann2023detection} have reported that C\&W are among the most challenging attacks to detect. C\&W attack leverages an optimization-based approach to craft adversarial examples, aiming to find the smallest perturbation necessary to cause misclassification by the target model. It minimizes both the magnitude of the perturbation and the distance to the neural network decision boundary. In the future, we will address these limitations by designing PDD ensembles to detect potential types of poisoned data.

\section{Conclusions and future work}\label{sec:Conclusions}
Self-supervised learning (SSL) enables scalable data curation, a cruial step for achieving generalization in machine learning models, particularly large VFMs. SSL meets the data demands of foundation models, however, the robustness of VFMs critically depends on the integrity of SSL-curated datasets used during training, as compromised training data can jeopardize the reliability of the subsequent inference phase. The defense mechanism proposed in this paper is highly scalable and actively detects poisoned data, thereby preserving the integrity of SSL-curated datasets used for foundation model training. We designed PDDs using a combination of the pretrained ImageBind model and four classifiers, namely, RF, KNN, NB, and SVM and evaluated PDDs using 176,200 images from three diverse datasets and three different adversarial attacks encompassing both ID and OOD scenarios. In the future, we will investigate a broader range of adversarial attacks relevant to data poisoning using PDD ensemble. Additionally, we plan to develop a one-class SVM for collective poisoned data detection.

\section*{Data Availability Statement}
Table~\ref{tab:Dataset} provides details of the various pre-trained models and datasets used in this paper and they are publicly available. 

\bibstyle{sn-mathphys-num}
\bibliography{references}

\end{document}